\documentclass[journal]{IEEEtran}

\usepackage{graphicx}
\usepackage{booktabs}
\usepackage{multirow}
\usepackage{amsmath,amssymb,amsfonts}
\usepackage{amsthm}
\usepackage{mathrsfs}
\usepackage{xcolor}
\usepackage{textcomp}
\usepackage{array}
\usepackage{tikz}
\usetikzlibrary{arrows.meta, positioning}
\usepackage{subcaption}
\usepackage{placeins}
\usepackage[authoryear,round]{natbib}
\bibpunct{(}{)}{;}{a}{}{,}
\usepackage{hyperref}
\hypersetup{colorlinks=true, linkcolor=black, citecolor=black, urlcolor=blue, pdfborder={0 0 0}}
\newcommand{\botrule}{\bottomrule}

\begin{document}

\title{A Persian Vision-Language-Action Dataset and Benchmark for Ambiguity- and Infeasibility-Aware Instruction-to-Plan Grounding}

\author{
\begin{minipage}[t]{0.31\textwidth}
\centering
Abdollah Pourasghari\\
Malek Ashtar University of\\Technology, Iran\\
\texttt{\footnotesize abdollah.pourasghari@iran.ir}
\end{minipage}%
\hfill
\begin{minipage}[t]{0.31\textwidth}
\centering
Saeedeh Sadat Sadidpour$^{*}$\\
Malek Ashtar University of\\Technology, Iran\\
\texttt{\footnotesize sadidpour@mut.ac.ir}
\end{minipage}%
\hfill
\begin{minipage}[t]{0.31\textwidth}
\centering
Mohammad Roustaei\\
Isfahan University of\\Technology, Isfahan, Iran\\
\texttt{\footnotesize roustaei.m@ec.iut.ac.ir}
\end{minipage}

\vspace{2pt}
{\footnotesize $^{*}$Corresponding author\par}
}

\markboth{Preprint}{Pourasghari \MakeLowercase{\textit{et al.}}: A Persian Vision-Language-Action Dataset and Benchmark}

\maketitle

\begin{abstract}
Converting natural-language instructions into executable action plans conditioned on a scene image is a core capability for embodied human-robot interaction, but existing benchmarks target English, simulated or video-grounded environments, and largely assume instructions are clear and executable. We introduce the first Persian vision-language-action dataset for this task: 258 instances over 230 real household/office photographs, annotated with reference plans, deliberate ambiguity and infeasibility labels, disambiguating question-answer pairs, and an auto-generated chain-of-thought rationale for training. Using this resource, we ablate retrieval-augmentation, chain-of-thought prompting, and an agentic disambiguation step, evaluated with an LLM-as-judge protocol and a new metric, PGOS-FA, extending prior work (PG2S) with step-order sensitivity, an over-generation penalty, and Persian-adapted embeddings. Paired statistical testing across 36 comparisons shows retrieval-augmentation is a large, robustly significant effect, while finer-grained additions are directionally consistent but not statistically distinguishable from noise at this scale, and the LLM-judge metric has near-zero power where PGOS-FA succeeds. A blind human-correlation study on 79 real outputs shows PGOS-FA tracks human judgment at least as closely as the LLM-judge. Testing generalization beyond the ablation's single backbone, we find that most of the gap between an open-weight and a substantially stronger closed planner is attributable to open-weight model generation rather than to the closed model's capability specifically, and that our best configuration significantly outperforms an independently reproduced external baseline despite a weaker backbone. The dataset, PGOS-FA implementation, and evaluation code will be made available upon reasonable request to support future work on instruction-to-plan grounding in low-resource languages.
\end{abstract}


\section{Introduction}\label{sec:intro}

Embodied agents that translate natural-language instructions into executable action plans are a core capability for human-robot interaction, assistive technology, and household automation. Recent progress in large language models (LLMs) and vision-language models (VLMs) has made single-image, language-conditioned planning increasingly practical: given a photograph of a scene and a free-form instruction, a model can be prompted to produce a step-by-step plan without task-specific training. This capability, however, has been developed and benchmarked almost entirely on English-language resources (ALFRED, DROID, EgoCOT --- Section~\ref{sec:relwork-benchmarks}), simulated or video-grounded environments, and instructions that are implicitly assumed to be clear and executable.

Two properties of real human instruction-giving are largely absent from this literature. First, \textbf{instructions are often genuinely ambiguous} relative to the scene --- multiple candidate referents, underspecified attributes --- requiring a system to recognize ambiguity and seek clarification rather than guess. Second, \textbf{instructions are sometimes not executable at all}, due to a physical or logical constraint visible in the scene, and a well-behaved system should recognize this rather than fabricate a plan. Neither property is a design focus of existing instruction-to-plan benchmarks, and neither has been studied for a low-resource language such as Persian, spoken by roughly 110 million people with comparatively limited NLP tooling and, to our knowledge, no prior vision-language-action resource of any kind.

This gap motivates the present work: we ask whether current LLM/VLM-based techniques --- retrieval-augmented generation, chain-of-thought prompting, and agentic disambiguation --- extend to a Persian, single-image, ambiguity-and-infeasibility-aware instruction-to-plan setting, and we build the resource needed to ask that question rigorously.

\textbf{Contributions.}
\begin{enumerate}
\item We introduce the first Persian vision-language-action dataset for instruction-to-action-plan grounding: 258 instances over 230 real (non-simulated) household/office photographs, with reference plans, explicit ambiguity and infeasibility annotations, disambiguating question-answer pairs, and an automatically-distilled chain-of-thought rationale for the training split (Section~\ref{sec:dataset}).
\item We conduct a controlled ablation of retrieval-augmentation, chain-of-thought prompting, and an agentic disambiguation step, and --- going beyond a standard ablation table --- quantify which of these effects are statistically distinguishable from noise at the dataset's current scale via paired significance testing across 36 comparisons (Section~\ref{sec:significance}), finding that retrieval-augmentation is the dominant, robustly significant effect, while finer-grained components remain directionally consistent but not yet statistically confirmed.
\item We introduce PGOS-FA, a Persian-adapted, order-sensitive, over-generation-penalizing extension of the PG2S automatic evaluation metric \citep{brienza2024multiagent}, and compare it against an LLM-as-judge protocol, showing PGOS-FA has substantially lower sampling variance on identical evaluation data (Section~\ref{sec:significance}). We further validate PGOS-FA against blind human ratings on 79 real model outputs, where it correlates with human judgment at least as closely as the LLM-judge (Section~\ref{sec:humancorr}).
\item We test generalization beyond the single open-weight backbone used throughout the ablation by decomposing the gap to a substantially stronger closed model into two isolated effects via a matched-generation control, showing that most of the gain is attributable to a newer open-weight model generation rather than to the closed model specifically (Section~\ref{sec:generalization}).
\item Using two independent lines of evidence --- paired significance testing at $n=39$ instances (Section~\ref{sec:significance}) and rank stability of all eight pipeline configurations under a second backbone (Section~\ref{sec:generalization}) --- we show that fine-grained ablation rankings are not robust, while the coarse-grained benefit of retrieval augmentation is, a finding we treat as a methodological contribution about benchmark practice at this scale rather than as a limitation.
\end{enumerate}

\section{Related Work}\label{sec:relwork}

\subsection{Prompting and Retrieval Techniques for LLM Reasoning}\label{sec:relwork-techniques}

We build on three well-established techniques rather than proposing new ones. Retrieval-augmented generation (RAG) \citep{lewis2020retrieval} grounds generation in retrieved exemplars rather than relying solely on parametric knowledge, and chain-of-thought (CoT) prompting \citep{wei2022chain,kojima2022large} elicits intermediate reasoning steps before a final answer. We also use an LLM-based agent for sub-task delegation (disambiguation), following the general framing of LLM agents as systems that perceive, reason about goals, and act \citep{luo2025large}. Our disambiguation agent (Section~\ref{sec:method-disambiguation}) is deliberately simple -- a prompted validator that asks at most one clarifying question per round for up to three rounds -- rather than adopting more principled recent approaches to when an agent should ask: \citet{liang2026roboclarify} clarify ambiguous home-robot instructions via a scenario-guided risk graph, while \citet{suri2026structured} formulate question selection itself as expected-value-of-information maximization over a structured space of candidate clarifications. Our contribution is not a new prompting or clarification technique but a rigorous, statistically-grounded empirical study of how well-established techniques compose for a specific, understudied task and language.

\subsection{Instruction-to-Action-Plan Benchmarks}\label{sec:relwork-benchmarks}

Table~\ref{tab:benchmark-comparison} compares our resource against the most closely related benchmarks along the dimensions relevant to our novelty claim, including the two very recent single-axis resources discussed below.

\begin{table*}[t]
\caption{Benchmark comparison. No prior resource we are aware of combines a low-resource target language, single-image (rather than simulated or video) grounding, and explicit annotation of both instruction ambiguity and physical infeasibility.}\label{tab:benchmark-comparison}
\centering
\footnotesize
\newcolumntype{R}[1]{>{\raggedright\arraybackslash}p{#1}}
\begin{tabular}{@{}R{0.09\textwidth}R{0.1\textwidth}R{0.17\textwidth}R{0.33\textwidth}R{0.1\textwidth}R{0.14\textwidth}@{}}
\toprule
Dataset & Language & Grounding & Scale & Ambig.\ labels & Infeas.\ labels \\
\midrule
ALFRED \citep{shridhar2020alfred} & English & Simulated (AI2-THOR) & $\sim$8{,}055 expert demos / $\sim$25{,}700 instructions & No & No \\
DROID \citep{khazatsky2025droid} & English & Real robot, video/trajectory & 76{,}000 episodes ($\sim$350h) & No & No \\
EgoCOT \citep{mu2023embodiedgpt,grauman2022ego4d} & English & Real, egocentric video & Large-scale (video-clip-derived) & No & No \\
RxR \citep{ku2020room} & English, Hindi, Telugu & Real building, simulated nav. & 126{,}000 instructions / 16{,}500 paths & No & No \\
AmbiK \citep{ivanova2025ambik} & English & Textual only (no image) & 1{,}000 pairs / 2{,}000 tasks & Yes (1{,}000) & No \\
VLN-NF \citep{su2026vlnnf} & English & Simulated nav.\ (Matterport3D) & Derived from REVERIE ($\sim$45\% retained after filtering) & No & Yes (feasible/NOT-FOUND pairs) \\
\textbf{Ours} & \textbf{Persian} & \textbf{Real, single static image} & \textbf{258 instances / 230 images} & \textbf{Yes (23)} & \textbf{Yes (13)} \\
\botrule
\end{tabular}
\normalsize
\end{table*}

ALFRED \citep{shridhar2020alfred} pairs $\sim$25K English instructions with expert demonstrations inside the AI2-THOR simulator, evaluated by task-success and goal-condition-success on the \emph{simulated} execution trace. DROID \citep{khazatsky2025droid} is a large-scale real-robot manipulation dataset with natural-language task descriptions, but is oriented toward low-level robot control trajectories rather than high-level step planning from a single image. EgoCOT \citep{mu2023embodiedgpt,grauman2022ego4d} auto-generates chain-of-thought step plans from egocentric video narrations (built on Ego4D) at scale, but is English-only and video-grounded.

None of these three resources (a) target Persian, (b) are grounded in a single static photograph (rather than a simulator or video), or (c) explicitly and deliberately label a subset of instructions as \emph{ambiguous} or \emph{physically infeasible} --- two failure modes that are central to real human-robot instruction-following but largely absent from benchmark design elsewhere. Our dataset is, to our knowledge, the first to combine all three properties for any language.

Multilingual coverage of embodied instruction-following remains sparse, with navigation the main exception. Room-Across-Room (RxR) \citep{ku2020room} is, to our knowledge, the most prominent example: a vision-and-language navigation dataset with dense, time-aligned instructions in English, Hindi, and Telugu --- two genuinely low-resource languages by NLP tooling standards --- collected natively in each language rather than by translation. RxR demonstrates that native-language, non-English embodied-instruction data collection is both feasible and valuable, but its task (step-by-step navigation instructions along a path in a simulated building) and evaluation (path-following success) differ substantially from single-image action-plan grounding, and it does not label instruction ambiguity or infeasibility. We are not aware of a published Persian-language counterpart to RxR, or of any single-image instruction-to-plan resource in a language other than English. A systematic 2026 audit of widely used vision-language-action corpora \citep{wanna2026limited} finds that even the English-language resources in this space rely on highly repetitive, template-like instruction language with limited structural variation. This is independent evidence that the field's language coverage problem is not only about which languages are represented, but how narrowly instructions are phrased even within English -- reinforcing the value of naturalistic, non-template instruction collection regardless of target language.

Two very recent, single-axis resources are worth situating relative to our joint ambiguity-and-infeasibility design. AmbiK \citep{ivanova2025ambik} contributes 1,000 ambiguous/unambiguous kitchen-task pairs with clarifying question-answer annotations at substantially larger scale than our ambiguous-instruction subset, but is fully textual -- it has no associated images, so ambiguity there is a property of the instruction text and a described scene, not something a system must resolve by looking. VLN-NF \citep{su2026vlnnf} addresses physical/logical infeasibility (via false-premise navigation goals) with an explicit NOT-FOUND action, directly paralleling our infeasibility labels, but in a simulated navigation setting rather than single-image action planning. Neither combines vision-grounded ambiguity and infeasibility together, and neither targets a low-resource language.

\subsection{LLM/VLM-Based Task Planning}\label{sec:relwork-planning}

A growing line of work uses LLMs to translate natural-language instructions into executable plans for embodied agents. SayCan \citep{ahn2022saycan} combines LLM-scored language-skill affordances with learned value functions to ground plans in what a robot can actually execute. LLM-Planner \citep{song2023llmplanner} uses hierarchical few-shot prompting with GPT-3 and a perception module, achieving competitive ALFRED performance with under 0.5\% of the training data. Multi-agent Visual Language Model Planning \citep{brienza2024multiagent} is the most directly related prior work: it plans from a single image plus instruction using GPT-4/GPT-4V multi-agent decomposition, and introduces the PG2S metric --- semantic-similarity-based automatic evaluation of predicted vs.\ reference plans --- which our proposed PGOS-FA metric (Section~\ref{sec:pgosfa}) directly extends and corrects for step-order sensitivity, over-generation penalty, and Persian-language embeddings. Other related systems include PLATO \citep{car2024plato} (affordance-driven planning without environment priors), P-RAG \citep{xu2024prag} (progressive retrieval for embodied task planning), and Wonderful Team \citep{wang2024wonderfulteam} (zero-shot multi-agent VLM planning with a supervisor/critic split). \citet{huang2024understanding} organize this space into a planning-strategy taxonomy -- task decomposition, multi-plan selection, external planners, reflection, and memory-augmented planning. Our ablation of RAG, CoT, and disambiguation maps onto that taxonomy's ``multi-plan selection + memory (retrieval) + reflection (disambiguation)'' categories, applied jointly for the first time to a single-image, ambiguity-aware, Persian-language setting.

\subsection{Automatic Evaluation of Generated Plans}\label{sec:relwork-evaluation}

Beyond PG2S (Section~\ref{sec:relwork-planning}, extended in Section~\ref{sec:pgosfa}), the closest evaluation paradigm is LLM-as-judge, increasingly used in place of surface-overlap metrics like BLEU/ROUGE for tasks requiring semantic and logical equivalence judgments rather than $n$-gram match. We directly compare an LLM-judge protocol (across seven candidate judge models and three few-shot regimes) against our PGOS-FA metric, and report both metrics' statistical power to detect true differences between system configurations (Section~\ref{sec:significance}).

\section{The Dataset}\label{sec:dataset}

\subsection{Motivation and Design Principles}\label{sec:dataset-motivation}

Existing benchmarks for grounding natural-language instructions into executable action sequences (ALFRED, DROID, EgoCOT) are built around English instructions and, in the case of ALFRED, a simulated household environment rather than real photographs. No prior resource targets Persian, nor does any prior resource explicitly encode two failure modes that are pervasive in real human-robot interaction: (i) instructions that are linguistically ambiguous given the scene, and (ii) instructions that are physically infeasible given the scene. We introduce, to our knowledge, the first Persian vision-language-action dataset that jointly covers real (non-simulated) household/office photographs, free-form Persian instructions, reference action plans, and explicit annotation of ambiguity and infeasibility.

\begin{figure*}[t]
\centering
\begin{subfigure}[t]{0.47\textwidth}
\centering
\includegraphics[width=\textwidth]{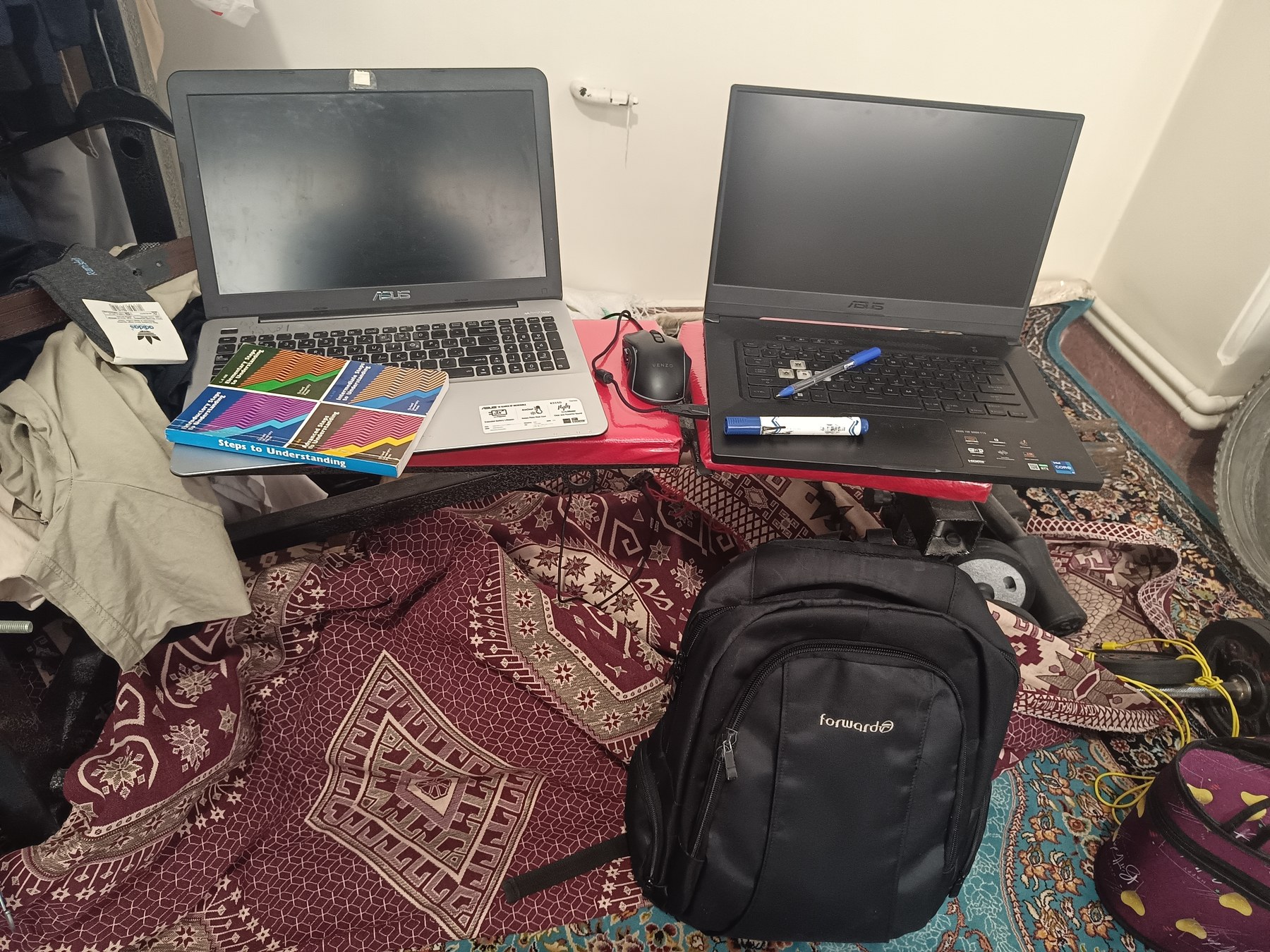}\\[2pt]
\includegraphics[width=\textwidth]{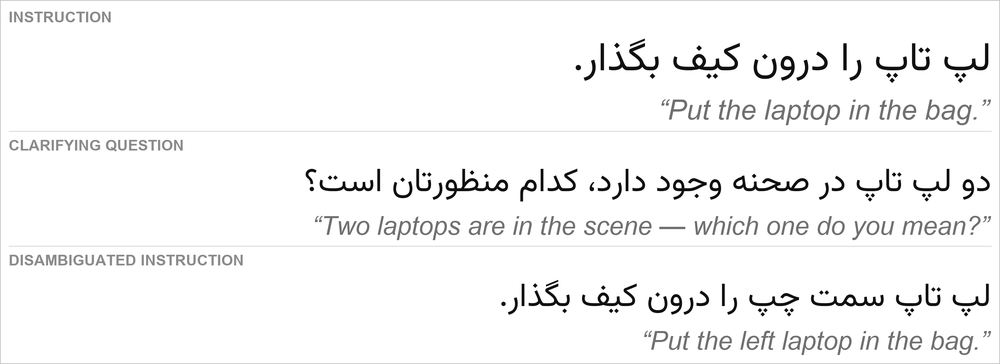}
\caption{Ambiguous instruction, original Persian with English gloss: the instruction, the disambiguation agent's clarifying question, and the resulting disambiguated instruction.}
\label{fig:ex-ambiguous}
\end{subfigure}
\hfill
\begin{subfigure}[t]{0.47\textwidth}
\centering
\includegraphics[width=\textwidth]{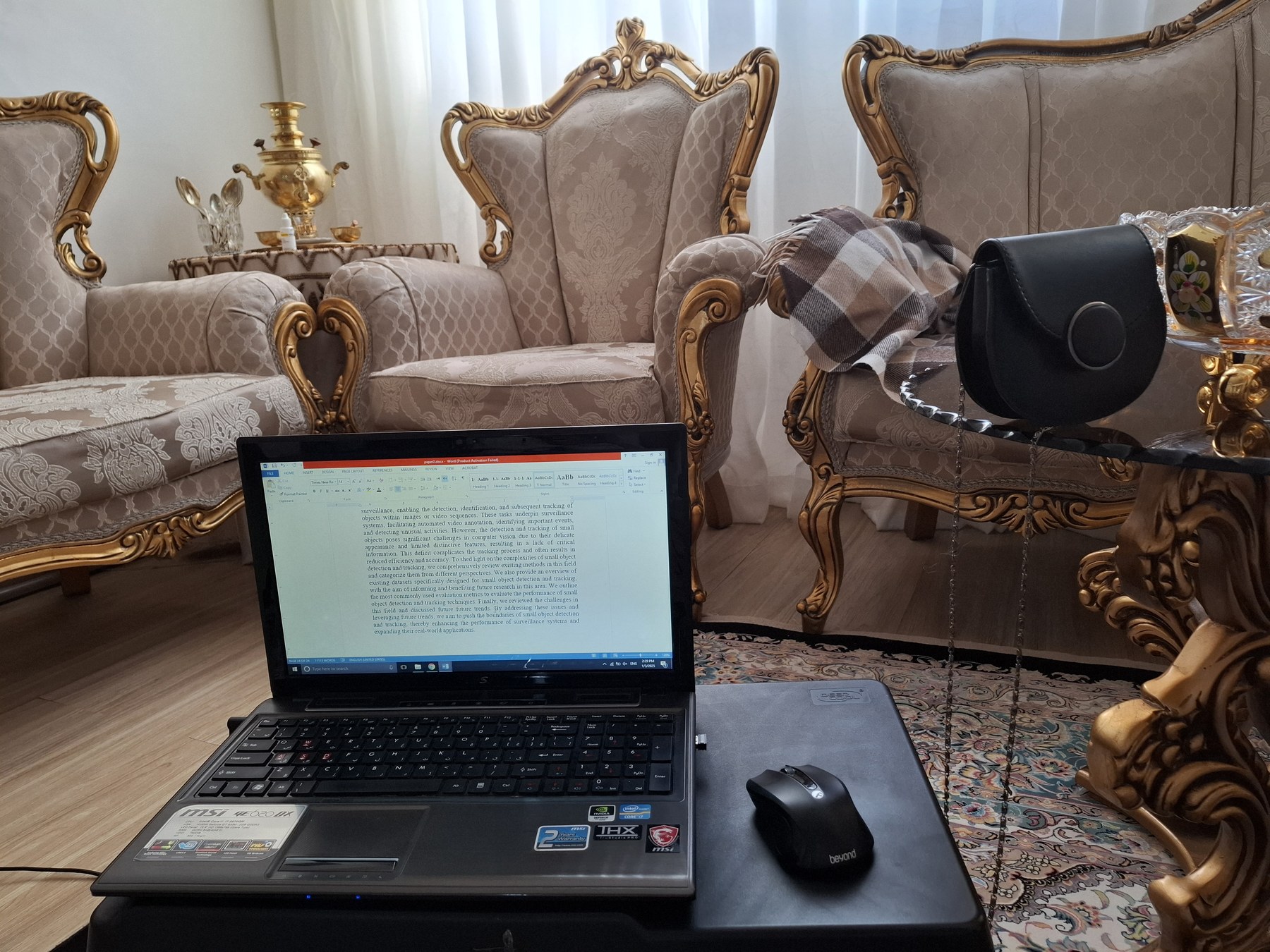}\\[2pt]
\includegraphics[width=\textwidth]{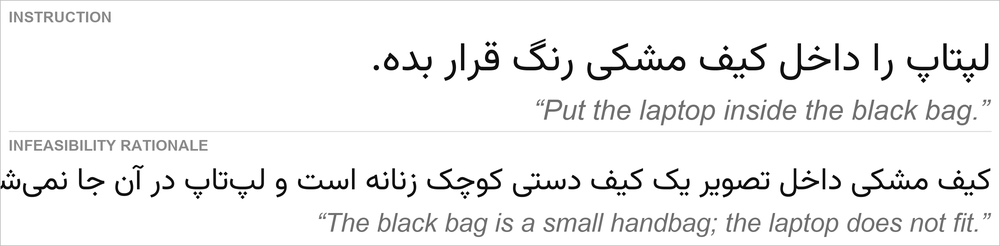}
\caption{Infeasible instruction, original Persian with English gloss: the instruction and the annotator-written infeasibility rationale.}
\label{fig:ex-infeasible}
\end{subfigure}
\caption{Two real instances from the dataset, illustrating the two deliberate-difficulty annotation types described above: (a) an ambiguous instruction requiring disambiguation, and (b) a physically infeasible instruction given the scene.}
\label{fig:dataset-examples}
\end{figure*}

Four design principles guided data collection:
\begin{itemize}
\item \emph{Single-image grounding}: each instance is grounded in exactly one static photograph, not a video or an interactive simulator, keeping the task tractable for single-pass VLM inference while still requiring genuine visual grounding.
\item \emph{Deliberate difficulty}: a subset of instructions is intentionally ambiguous or infeasible (examples in Figure~\ref{fig:dataset-examples}), so a system cannot succeed by pattern-matching template instructions.
\item \emph{Cultural and linguistic authenticity}: scenes, objects, and instructions are drawn from everyday Iranian home/office life rather than translated from an English source.
\item \emph{Distillation-ready supervision}: a chain-of-thought rationale column is auto-generated for the training split, enabling future fine-tuning of smaller open-weight models on this resource.
\end{itemize}

\subsection{Data Collection}\label{sec:dataset-collection}

Data collection involved 19 contributors, each producing instances consisting of (a) a photograph of a real home or office scene, (b) a natural-language Persian instruction referencing that scene, and (c) a proposed reference sequence of executable steps. Contributors worked from a written task description specifying these three deliverables and the general goal of covering varied, realistic home/office scenarios; photographs were predominantly taken in domestic settings, with a smaller proportion from office environments, to maximize object and layout diversity. We did not employ a formal annotator training session or a pilot-and-calibrate round before full-scale collection, and the original task description is not currently available as a standalone document for release alongside the dataset -- we list this as an explicit annotation-methodology limitation in Section~\ref{sec:limitations} rather than understate it.

\subsection{Curation, Quality Control, and Schema}\label{sec:dataset-schema}

Raw contributions underwent multi-stage curation performed by the research team, not the original 19 contributors:
\begin{enumerate}
\item selection of higher-quality instances from the raw pool;
\item review and, where needed, correction or replacement of instructions to ensure they were well-formed and clearly grounded in the associated image;
\item deliberate injection of ambiguity into a subset of instructions that were originally unambiguous, specifically to raise task difficulty (Section~\ref{sec:dataset-motivation});
\item review of reference action sequences for technical correctness (physical plausibility given the scene) and stylistic consistency across contributors;
\item addition of the derived annotation fields -- infeasibility rationale, clarifying question-answer pairs, and the disambiguated instruction rewrite (the last produced with AI assistance conditioned on the human-written question-answer pair, then reviewed by the curation team).
\end{enumerate}
This curation was performed by a small team without a second, independent pass over the same instances. No inter-annotator agreement statistic is therefore available for either the original contributions or the curation decisions themselves; Section~\ref{sec:limitations} discusses the consequences of this for the reliability claims this paper can make about the reference labels.

Table~\ref{tab:schema} lists the fields of each finalized instance.

\begin{table*}[t]
\centering
\caption{Dataset schema.}\label{tab:schema}
\begin{tabular}{@{}p{0.27\textwidth}p{0.61\textwidth}@{}}
\toprule
Field & Description \\
\midrule
\texttt{image} & Filename of the associated photograph \\
\texttt{instruction} & The raw Persian instruction as given \\
\texttt{actions} & Reference action plan (numbered steps) \\
\texttt{\footnotesize executable\_or\_\allowbreak notexecutable} & Binary flag: is the instruction physically/logically executable given the scene? \\
\texttt{clarification\_\allowbreak question} & Disambiguating question(s) and answer(s), present only when the instruction is ambiguous \\
\texttt{Clear\_Instruction} & Disambiguated rewrite of the instruction, derived from the clarification Q\&A \\
\texttt{COT} (training split only) & Auto-generated chain-of-thought rationale (scene description $\to$ object/relation graph $\to$ reasoning $\to$ derivation of the reference plan) \\
\botrule
\end{tabular}
\end{table*}

\subsection{Statistics}\label{sec:dataset-statistics}

The dataset comprises 230 unique photographs and 258 total instances: a training split of 159 instances (120 used directly for retrieval, 39 held out for ablation/validation) and a test split of 99 instances. Thirteen instances are flagged infeasible and 23 are flagged ambiguous (one instance is both). Table~\ref{tab:split-breakdown} breaks this down per split.

\begin{table}[t]
\caption{Instruction type by split. The Ambiguous and Infeasible columns each include the single instance that is both, so Plain + Ambiguous + Infeasible $-$ Both = Total. The Retrieval-pool and Validation rows together make up the 159-instance training split described above.}\label{tab:split-breakdown}
\begin{tabular}{@{}lccccc@{}}
\toprule
Split & Total & Plain & Ambiguous & Infeasible & Both \\
\midrule
Retrieval pool & 120 & 98 & 17 & 6 & 1 \\
Validation & 39 & 37 & 0 & 2 & 0 \\
Test & 99 & 88 & 6 & 5 & 0 \\
\textbf{All} & \textbf{258} & \textbf{223} & \textbf{23} & \textbf{13} & \textbf{1} \\
\botrule
\end{tabular}
\end{table}

Infeasible instructions are spread evenly (5\% of each split), but ambiguous instructions are not: 14\% of the retrieval pool is ambiguous, 6\% of the test split, and none of the 39 validation instances. This matters for interpreting the ablation (Section~\ref{sec:ablation-results}), which runs on the validation split: there, the disambiguation agent's intended benefit cannot be tested on a genuinely ambiguous instruction at all.

\subsection{Automatic Chain-of-Thought Generation}\label{sec:dataset-cot}

To support future fine-tuning of smaller open models on this resource, we augment the training split with an automatically generated chain-of-thought rationale for each instance. Three instances were manually annotated with expert human reasoning to serve as in-context exemplars; a stronger model was then prompted, given the instruction, image, reference plan, and the three exemplars, to produce a four-part rationale for every remaining training instance: (1)~a detailed scene description, (2)~a structured object/relation graph (JSON, subject--relation--object triples), (3)~explicit reasoning about obstacles, prerequisite actions, and physical/logical feasibility, and (4)~the derivation of the final action plan, aligned with the human-written reference. This design lets the CoT column double as both a training signal for distillation and a source of few-shot exemplars for CoT prompting at inference time (Section~\ref{sec:method}).

\subsection{Known Limitations of the Resource}\label{sec:dataset-limitations}

Two limitations of the dataset itself --- its scale, and the lack of independent second-annotator verification of reference plans and labels --- are discussed in full, alongside their consequences for this paper's claims, in Section~\ref{sec:limitations}. We flag one further point here: instructions are Persian-only, and the English/Persian comparison in Section~\ref{sec:results} concerns the \emph{prompt} language used to query models, not translated data.

\section{Method}\label{sec:method}

\subsection{Task Definition}\label{sec:method-task}

Given an instruction $I$ in Persian and a single RGB image $V$ of a static scene, the task is to produce a sequence of executable steps $P = (p_1, \ldots, p_n)$ that accomplishes $I$ in the scene depicted by $V$, or to determine that $I$ is not executable (due to a physical or logical constraint visible in $V$) and state why. A reference plan $G = (g_1, \ldots, g_m)$ (or an infeasibility judgment) is available for every instance in the dataset (Section~\ref{sec:dataset}).

\subsection{Pipeline Architecture}\label{sec:method-pipeline}

\begin{figure*}[t]
\centering
\resizebox{0.85\textwidth}{!}{%
\begin{tikzpicture}[
  box/.style={draw, rounded corners, align=center, minimum height=1.8cm, text width=2.8cm, font=\normalsize, inner sep=6pt},
  optbox/.style={box, dashed},
  lbl/.style={font=\footnotesize, align=center, text width=2.8cm},
  arr/.style={-{Latex[length=3mm]}, thick},
  node distance=1.2cm
]
\node[box] (input) {Image $V$ +\\ Instruction $I$};
\node[optbox, right=of input] (disambig) {Disambiguation Agent\\ (validator, human-sim,\\ merger; $\le3$ rounds)};
\node[optbox, right=of disambig] (retrieval) {Retrieval\\ ($k{=}3$ exemplars,\\ Tooka-SBERT)};
\node[box, right=of retrieval] (planner) {Planner\\ (+ Chain-of-Thought)};
\node[box, right=of planner] (output) {Plan $P$ or\\ ``not executable''};
\draw[arr] (input) -- (disambig);
\draw[arr] (disambig) -- (retrieval);
\draw[arr] (retrieval) -- (planner);
\draw[arr] (planner) -- (output);
\node[lbl, below=0.2cm of disambig] {configurations 6--8 only};
\node[lbl, below=0.2cm of retrieval] {configurations 2--8 only};
\node[lbl, below=0.2cm of planner] {CoT in configurations 4--8};
\end{tikzpicture}%
}
\caption{Pipeline architecture. Every configuration passes the (possibly disambiguated) instruction and image to the planner; the disambiguation agent and the retrieval step (dashed boxes) are each present only in a subset of the eight configurations, and the planner uses chain-of-thought reasoning only in a further subset -- exactly which subset is specified per configuration in Table~\ref{tab:configurations}.}
\label{fig:pipeline}
\end{figure*}
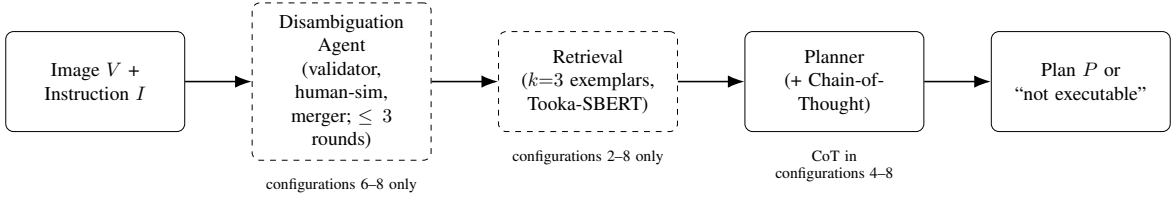

We study eight pipeline configurations, each layering one additional component on the previous, following an ablation design (Table~\ref{tab:configurations}; architecture in Figure~\ref{fig:pipeline}). All configurations use the same open-weight VLM (Gemma-3-27B-IT) for every agent role; Section~\ref{sec:generalization} tests whether the conclusions are backbone-specific by re-running configurations under a newer open-weight model and under a substantially stronger closed planner.

\begin{table*}[t]
\centering
\caption{Pipeline configurations.}\label{tab:configurations}
\begin{tabular}{@{}cp{0.3\textwidth}cp{0.22\textwidth}cc@{}}
\toprule
\# & Configuration & RAG ex. & Example content & CoT & Disambig. \\
\midrule
1 & Baseline & 0 & --- & No & No \\
2 & RAG (1-shot) & 1 & reference plan & No & No \\
3 & RAG (3-shot) & 3 & reference plan & No & No \\
4 & RAG (3-shot) + CoT & 3 & reference plan & Yes & No \\
5 & RAG (3-shot, unclarified ex.) + CoT & 3 & reference plan, unclarified instruction & Yes & No \\
6 & RAG (3-shot) + Disambiguation & 3 & reference plan & No & Yes \\
7 & RAG (3-shot) + Disambiguation + CoT & 3 & reference plan & Yes & Yes \\
8 & RAG (3-shot, CoT ex.) + Disambiguation + CoT & 3 & reference plan + CoT rationale & Yes & Yes \\
\botrule
\end{tabular}
\end{table*}

Configuration 2 retrieves the single most similar training instance by cosine similarity over Tooka-SBERT instruction embeddings and injects it, with its reference plan, into the prompt; configuration 3 retrieves three. Configuration 4 keeps the same retrieved reference-plan exemplars but switches the planner to a chain-of-thought prompt that asks it to describe the scene, extract an object/relation graph, and reason about obstacles and feasibility before emitting the final plan. Configuration 5 tests whether keeping the retrieved examples' instructions in their original, unclarified form (rather than their disambiguated rewrite) changes performance. Configurations 6--8 add the disambiguation agent (Section~\ref{sec:method-disambiguation}) before planning, with configuration 8 additionally using CoT-annotated retrieved examples on top of configuration 7. Every configuration is run with both a Persian and an English system prompt.

\textbf{Retrieval and train/test separation.} For the validation ablation (Section~\ref{sec:ablation-results}), every retrieving configuration's index contains only the 120-instance retrieval pool, so the 39 validation instances are never retrievable during their own evaluation; the 1-shot configuration's index was additionally restricted to the executable subset of that pool. Configurations 7 and 8, the only configurations carried forward to held-out test evaluation (Table~\ref{tab:testset}), instead retrieve from an index containing the full 159-instance training split (Section~\ref{sec:dataset-statistics}) -- the validation instances are legitimately part of the retrieval pool at test time, since they are training data from the test set's perspective. Retrieved exemplars are injected as text only -- instruction, reference plan and, where applicable, CoT rationale -- never as images. Because photographs are reused across instructions, 13 test photographs also appear in training instances with different instructions; this partial overlap (same scene, different instruction) is retained throughout.

\subsubsection{Disambiguation Agent}\label{sec:method-disambiguation}

A three-stage interactive procedure (max.\ three rounds) determines whether $I$ is ambiguous given $V$: (1)~a validator agent analyzes the scene and either confirms the instruction is clear or emits a clarifying question; (2)~if a question is raised, a \emph{human-simulator} agent (a separate LLM call, standing in for the end user during automated evaluation) answers using the instance's ground-truth clarification Q\&A, or responds ``don't know'' if no matching ground-truth answer exists; (3)~an instruction-merger agent rewrites $I$ into an unambiguous instruction $I'$ given the question-answer pair. $I'$ (or $I$, if never disambiguated) is then passed to the planner. This three-round limit is the design used for configuration 8 and for every generalization and rank-stability experiment in this paper (Section~\ref{sec:generalization}). Configurations 6 and 7's original ablation results (Table~\ref{tab:ablation}) were instead produced with a two-round limit, an inconsistency between the original notebooks rather than a deliberate design difference; we standardize all subsequent experiments to three rounds. Section~\ref{sec:qualitative} shows that the automated human-simulator's inability to answer novel clarifying questions is itself a notable source of downstream error, distinct from the planner's own capability.

\subsection{Automatic Evaluation}\label{sec:method-evaluation}

Two complementary automatic metrics are used throughout, motivated by the impracticality of human evaluation at ablation scale and the limitations of surface-overlap metrics (BLEU/ROUGE) for this task.

\subsubsection{LLM-as-Judge}\label{sec:llmjudge}

A candidate judge model receives $I$, $V$, $G$, and $P$, and returns a binary correctness verdict. Seven candidate judges (GPT-4.1-nano/mini/4.1, GPT-4o/4o-mini, o4-mini, Gemini-2.5-Flash) were compared across three regimes (zero-shot, few-shot text-only, few-shot text+image) on a 16-instance controlled set (8 correct / 8 incorrect reference-vs.-generated pairs); Gemini-2.5-Flash with CoT-augmented few-shot exemplars was selected as the production judge (0.88 accuracy on the controlled set).

\subsubsection{PGOS-FA}\label{sec:pgosfa}

PGOS-FA (Persian Goal-Order-Sentence, Fixed \& Adapted) extends PG2S \citep{brienza2024multiagent} to address four limitations of PG2S in this setting: order-insensitivity, a fixed similarity threshold, no penalty for over-generation, and embedding and parsing tools that do not support Persian. It combines two sub-scores, each computed with an explicit order component.

\textbf{Infeasibility short-circuit.} Instances whose reference plan marks the instruction as not executable are not scored by the similarity procedure below at all: if the generated output also declares infeasibility, the instance scores $1$; if the ground truth says infeasible but the model produced a plan anyway, or vice versa, the instance scores $0$. Only instances where both reference and prediction agree the instruction is feasible proceed to the similarity computation.

\textbf{Goal-wise similarity.} Each plan step is decomposed by dependency parsing, using Stanza's \citep{qi2020stanza} Persian pipeline (trained on the Persian Dependency Treebank), into a $(\text{verb}, \text{noun}_1, \ldots, \text{noun}_k)$ tuple (verb plus its subject/object dependents). For reference step $i$ with tuple $(v_i, n_{i,1}, \ldots, n_{i,k})$ and predicted step $j$ with $(v_j, n_{j,1}, \ldots, n_{j,l})$, using Tooka-SBERT embeddings $e(\cdot)$:
\begin{equation}
\sigma_{verb}(i,j) = \cos\bigl(e(v_i), e(v_j)\bigr)
\end{equation}
\begin{equation}
\resizebox{\linewidth}{!}{$\sigma_{noun}(i,j) = \begin{cases} \dfrac{1}{\min(k,l)}\displaystyle\sum_{t=1}^{\min(k,l)} \cos\bigl(e(n_{i,t}), e(n_{j,t})\bigr) & \min(k,l) > 0 \\[2.2ex] 0 & \min(k,l) = 0 \end{cases}$}
\end{equation}
\begin{equation}
sim_{goal}(i,j) = \sigma_{verb}(i,j) \cdot \sigma_{noun}(i,j)
\end{equation}

Reference steps are greedily matched to their highest-similarity unmatched predicted step (each predicted step used at most once); unmatched reference steps (predicted plan too short) \emph{and} leftover unmatched predicted steps (over-generation) each contribute a $0$ to the score list --- this explicit over-generation penalty is what PG2S lacked. Let $\mathcal{M} = \{(i, \pi(i))\}$ be the resulting set of matched index pairs, in reference order, and let $m,n$ be the reference/predicted plan lengths. Content score:
\begin{equation}
S_{content}^{goal} = \frac{1}{\max(m,n)}\sum_{(i,\pi(i)) \in \mathcal{M}} sim_{goal}\bigl(i, \pi(i)\bigr)
\end{equation}
Order score via Kendall's $\tau$ \citep{kendall1938new} over the matched-pair index sequence, normalized to $[0,1]$:
\begin{equation}
S_{order}^{goal} = \begin{cases} \dfrac{\tau\bigl(\{i\}, \{\pi(i)\}\bigr) + 1}{2} & |\mathcal{M}| > 1 \\[1.8ex] 0 & |\mathcal{M}| \le 1 \end{cases}
\end{equation}
\begin{equation}
Score_{goal} = (1-\alpha) \cdot S_{content}^{goal} + \alpha \cdot S_{order}^{goal}
\end{equation}
Kendall's $\tau$ is undefined for fewer than two matched pairs, a common case for the many short (one- or two-step) plans in this dataset; the implementation used throughout this paper sets $S_{order}$ to $0$ rather than treating order as trivially satisfied in this case. This is a conservative but non-obvious convention -- a correctly matched single-step plan receives no order credit -- that applies identically to $S_{order}^{sentence}$ below, and we document it here for reproducibility.

\textbf{Sentence-wise similarity.} Identical matching/order procedure and, in our implementation, the same verb/noun-decomposed tuples as the goal-wise term above are re-embedded and compared as whole units with Tooka-SBERT rather than PG2S's MPNet (not Persian-trained); the two sub-scores therefore differ in \emph{how} similarity between two steps is computed (compositional verb$\times$noun product vs.\ a single whole-tuple embedding comparison) rather than in what linguistic representation of the step they start from:
\begin{equation}
Score_{sentence} = (1-\alpha) \cdot S_{content}^{sentence} + \alpha \cdot S_{order}^{sentence}
\end{equation}

A small number of instances contain at least one step whose verb has two or more object/subject dependents (e.g., ``put the bread and the milk in the fridge''); the sentence-wise embedding call does not support this multi-dependent case and raises an error rather than returning a score. We exclude such instances from the affected table's mean and any paired test built on it, rather than silently assigning a default score; the reduced $n$ is reported explicitly wherever this applies (Tables~\ref{tab:generalization} and \ref{tab:externalbaseline}).

\textbf{Final score.}
\begin{equation}
PGOS\text{-}FA = \frac{Score_{goal} + Score_{sentence}}{2} \in [0,1]
\end{equation}

$\alpha$ (order-weight; $\alpha=0.3$ in the implementation that produced every reported score in this paper) controls the relative importance of step ordering versus content match, addressing PG2S's complete disregard for step order. Unlike PG2S, which binarized each matched pair against a fixed similarity threshold (0.708 for goal-wise matching, 0.675 for sentence-wise matching) before averaging, PGOS-FA uses the continuous similarity score directly, removing a hyperparameter whose optimal value is itself embedding-model- and language-dependent.

\paragraph{Does each component earn its place?} We ablate PGOS-FA's own components on the 16-instance correct/incorrect controlled set (Section~\ref{sec:llmjudge} background), comparing the full metric against four variants: \emph{content\_only} ($\alpha=0$, order term removed), \emph{order\_only} ($\alpha=1$, content term removed), \emph{goal\_only} (goal-wise sub-score alone), and \emph{sentence\_only} (sentence-wise sub-score alone). Table~\ref{tab:pgosfa-ablation} reports the mean score for correct and incorrect plans under each variant and Cohen's $d$ as an effect-size measure of how well each variant separates the two groups.

\begin{table*}[t]
\centering
\caption{PGOS-FA component ablation on the 16-instance controlled set (8 correct / 8 incorrect plans).}\label{tab:pgosfa-ablation}
\begin{tabular}{@{}lcccc@{}}
\toprule
Variant & Mean (correct) & Mean (incorrect) & Gap & Cohen's $d$ \\
\midrule
Full (official) & 0.883 & 0.595 & 0.287 & 1.33 \\
Content only ($\alpha=0$) & 0.876 & 0.513 & 0.363 & 1.52 \\
Order only ($\alpha=1$) & 0.899 & 0.788 & 0.111 & 0.53 \\
Goal-wise only & 0.866 & 0.566 & 0.300 & 1.29 \\
Sentence-wise only & 0.899 & 0.624 & 0.275 & 1.37 \\
\botrule
\end{tabular}
\end{table*}

Two findings temper an unqualified claim that adding the order term improves the metric. First, the order-only variant has markedly weaker discriminative power ($d=0.53$) than every content-based variant ($d>1.2$) on this set, and content-only slightly \emph{outperforms} the official full metric ($d=1.52$ vs.\ $1.33$); part of this gap is likely the $|\mathcal{M}|\le 1$ convention noted above, which zeroes the order term outright on the set's many short plans regardless of whether order is actually violated. Second, both the goal-wise-only and sentence-wise-only variants perform comparably to the full metric on their own, suggesting some redundancy between the two sub-scores on this small controlled set. We do not read this as evidence that step order is unimportant to plan correctness in general: the 16 controlled instances were constructed to contrast semantically correct and incorrect plans, and the ``incorrect'' plans in this set are predominantly content errors (wrong object, missing step, wrong outcome) rather than pure step-reordering errors, so this particular test is not well-powered to showcase the order term's intended benefit.

A metrics-focused reader should not have to take the value of the order term on faith, so we constructed the dedicated test this diagnosis calls for: a controlled set contrasting plans that differ \emph{only} in step order. For every reference plan in the dataset with at least three steps (239 plans; 233 scored successfully, 6 excluded by the scoring failure described above), we compare the reference against two variants built from its own steps: the reference itself (content and order both correct) and the same steps in a randomly permuted, guaranteed-different order (content correct, order violated). Because the two variants share identical step content by construction, only the order sub-score can differ between them. It does, sharply: mean order-only score $1.000$ for order-correct plans versus $0.531$ for order-violated plans (Cohen's $d = 3.15$; Wilcoxon signed-rank $p < 10^{-38}$, $n=233$) -- a far larger effect than either the confounded order-only result above or the full metric's content-discrimination result, because this test isolates exactly the one thing it is measuring. We retain the order term at $\alpha=0.3$ throughout this paper on the strength of this result: the order term demonstrably does what it is designed to do, the full metric still separates correct from incorrect plans with a large effect size ($d=1.33$) on the original controlled set, and, more directly relevant to real model outputs, it is this exact configuration that Section~\ref{sec:humancorr} validates against blind human judgment.

\section{Results and Analysis}\label{sec:results}

\subsection{Ablation Results}\label{sec:ablation-results}

Table~\ref{tab:ablation} reports mean PGOS-FA and LLM-judge accuracy for all eight configurations in both languages, with 95\% confidence intervals from a nonparametric bootstrap (1{,}000 resamples with replacement, percentile method \citep{efron1979bootstrap}) over the $n=39$ validation instances. Two configurations are carried forward to held-out test evaluation ($n=99$, Table~\ref{tab:testset}): the highest-PGOS-FA configuration (configuration~8, RAG with CoT-annotated examples + Disambiguation + CoT, English prompt) and the highest-LLM-judge-accuracy configuration (configuration~7, RAG + Disambiguation + CoT, English prompt). On the held-out set their PGOS-FA order reverses, so configuration~7 is the best held-out configuration and the one used in the rest of this section.

\begin{table*}[t]
\centering
\caption{Ablation results, $n=39$ validation instances. Configuration numbering matches Table~\ref{tab:configurations}. Bold marks the per-metric maximum at validation scale; Section~\ref{sec:significance} quantifies why several of these apparent differences should not be read as proven effects without the significance results that follow.}\label{tab:ablation}
\newcolumntype{C}[1]{>{\centering\arraybackslash}p{#1}}
\begin{tabular}{@{}p{0.27\textwidth}cC{0.23\textwidth}C{0.23\textwidth}@{}}
\toprule
Configuration & Lang & PGOS-FA & LLM-judge acc.\ \\
 & & (95\% CI) & (95\% CI) \\
\midrule
1. Baseline & En & 0.341 [0.277, 0.412] & 0.205 [0.077, 0.333] \\
1. Baseline & Fa & 0.369 [0.306, 0.438] & 0.231 [0.103, 0.385] \\
2. RAG (1-shot) & En & 0.516 [0.424, 0.604] & 0.256 [0.128, 0.385] \\
2. RAG (1-shot) & Fa & 0.540 [0.453, 0.628] & 0.333 [0.179, 0.487] \\
3. RAG (3-shot) & En & 0.535 [0.452, 0.619] & 0.359 [0.231, 0.513] \\
3. RAG (3-shot) & Fa & 0.527 [0.456, 0.606] & 0.333 [0.179, 0.487] \\
4. RAG (3-shot) + CoT & En & 0.524 [0.446, 0.603] & 0.308 [0.179, 0.462] \\
4. RAG (3-shot) + CoT & Fa & 0.516 [0.450, 0.591] & 0.308 [0.179, 0.462] \\
5. RAG (unclarified) + CoT & En & 0.491 [0.412, 0.573] & 0.282 [0.154, 0.436] \\
5. RAG (unclarified) + CoT & Fa & 0.547 [0.474, 0.619] & 0.410 [0.256, 0.564] \\
6. RAG + Disambiguation & En & 0.554 [0.477, 0.635] & 0.359 [0.205, 0.513] \\
6. RAG + Disambiguation & Fa & 0.526 [0.449, 0.604] & 0.333 [0.179, 0.462] \\
7. RAG + Disambiguation + CoT & En & 0.548 [0.472, 0.629] & \textbf{0.462} [0.308, 0.615] \\
7. RAG + Disambiguation + CoT & Fa & 0.554 [0.484, 0.630] & 0.410 [0.256, 0.564] \\
8. RAG (CoT ex.) + Disambig.\ + CoT & En & \textbf{0.563} [0.483, 0.647] & 0.410 [0.256, 0.564] \\
8. RAG (CoT ex.) + Disambig.\ + CoT & Fa & 0.533 [0.463, 0.613] & 0.359 [0.205, 0.513] \\
\botrule
\end{tabular}
\end{table*}

\begin{table*}[t]
\centering
\caption{Held-out test set ($n=99$). CIs use the same bootstrap procedure as Table~\ref{tab:ablation}.}\label{tab:testset}
\begin{tabular}{@{}p{0.4\textwidth}p{0.22\textwidth}p{0.22\textwidth}@{}}
\toprule
Configuration & PGOS-FA & LLM-judge accuracy \\
\midrule
Baseline (English) & 0.376 [0.328, 0.427] & 0.212 [0.131, 0.293] \\
RAG + Disambiguation + CoT (English) & 0.524 [0.478, 0.571] & 0.253 [0.172, 0.343] \\
RAG (CoT ex.) + Disambiguation + CoT (English) & 0.507 [0.458, 0.566] & 0.283 [0.192, 0.364] \\
\botrule
\end{tabular}
\end{table*}

\subsection{Statistical Significance of Ablation Effects}\label{sec:significance}

Bootstrap confidence intervals alone leave open whether apparent differences between configurations reflect a true effect or sampling noise. We therefore ran paired significance tests (Wilcoxon signed-rank for PGOS-FA's continuous scores, exact McNemar for the LLM-judge's binary verdicts) across 36 comparisons spanning four hypothesis families --- each configuration vs.\ baseline, the main incremental chain, targeted component-ablation branches (CoT alone, example clarity), and the English-vs-Persian prompt effect at every configuration level --- with Holm--Bonferroni correction within each family. We scope correction to within each family, rather than across the whole paper's confirmatory tests, following standard practice for planned, conceptually distinct hypothesis families \citep[e.g.,][]{bender2001adjusting}; Section~\ref{sec:externalbaseline} reports an additional, more conservative pooled check across all of this paper's confirmatory PGOS-FA tests.

\textbf{Finding 1 --- retrieval is the dominant, statistically bulletproof effect.} All 14 ``any configuration with retrieval vs.\ baseline'' comparisons are significant on PGOS-FA after correction (largest corrected $p \approx 0.0017$; several $p < 0.0001$). Simply adding even a single retrieved exemplar produces the majority of the total improvement over baseline seen anywhere in the ablation.

\textbf{Finding 2 --- no genuinely new downstream component reaches significance once retrieval is present, at $n=39$.} Of 22 comparisons testing finer-grained additions (more exemplars, disambiguation, CoT, or prompt language; the incremental chain plus the targeted branches), 20 do not survive correction; the remaining two are the incremental chain's first link (baseline $\to$ one retrieved exemplar, one per language) --- mechanically the same retrieval-vs-baseline comparison already reported as significant in Finding 1, not a novel finer-grained result. This directly qualifies an incremental narrative in which each added component beyond the first retrieved exemplar is treated as independently proven to help. Most additions move PGOS-FA in the expected direction; the two exceptions are small drops when CoT is added to 3-shot retrieval, and when CoT is added on top of disambiguation under the English prompt. Several individual comparisons are suggestive ($p < 0.1$ uncorrected for a few), but the current sample size cannot statistically distinguish any of them from a null effect once retrieval is already in the pipeline. The validation split also contains no ambiguous instructions (Section~\ref{sec:dataset-statistics}), so these comparisons cannot test disambiguation's intended benefit at all; they can only show that it does not measurably harm clear instructions.

\textbf{Finding 3 --- the LLM-judge metric has near-zero power to distinguish configurations at this scale.} Zero of 36 comparisons are significant on the LLM-judge outcome, including baseline-vs-best-configuration on the $n=99$ held-out set ($p=0.42$ and $p=0.25$ for the two carried-forward configurations, respectively --- the headline accuracy gains of $0.212 \to 0.253/0.283$ in Table~\ref{tab:testset} are consistent with noise at this sample size). PGOS-FA, by contrast, detects the same baseline-vs-best gap with $p < 10^{-7}$ on the identical instances, providing quantitative support for PGOS-FA's design goal of lower sampling variance.

\textbf{Implication.} We report Findings 1--3 rather than a per-component narrative in which each addition is independently proven. Framing a $n=39$-validated, largely non-significant sequence of component additions as a chain of proven individual contributions would overstate what the data support. The clear, well-powered finding --- retrieval-augmentation from this dataset is highly effective, and PGOS-FA reliably detects it --- is a stronger and more defensible empirical claim. It also directly motivates our call in Section~\ref{sec:limitations} for a larger benchmark: distinguishing the finer-grained component effects requires more data than a 39-instance validation split can provide. We view this transparency as a methodological strength rather than a shortfall: a paper that reports only the effects that happen to be significant, without explicitly testing power, risks a form of overclaiming that a dataset of this scale makes especially easy to fall into.

\subsection{Human Correlation Validation of PGOS-FA}\label{sec:humancorr}

Section~\ref{sec:pgosfa}'s controlled-set validation establishes that PGOS-FA separates correct from incorrect plans on 16 hand-constructed instances; it does not by itself establish that PGOS-FA tracks human judgment on real model outputs across the full range of quality produced in practice. We address this directly with a blind human-correlation study.

\textbf{Protocol.} We drew a stratified sample of 79 (image, instruction, reference plan, generated plan) tuples from the pooled outputs of the three configurations evaluated on the held-out test set (the baseline and the two configurations in Table~\ref{tab:testset}), binned into five quantiles of PGOS-FA score so the sample spans the full quality range rather than clustering near the best configuration's ceiling; this covers three of the eight ablation configurations rather than all eight. Three independent annotators, blind to configuration identity and to both automatic scores, rated each generated plan's agreement with the reference plan on a 1--5 Likert scale using a dedicated annotation tool, working independently, in a fixed instance order, without consulting one another.

\textbf{Inter-annotator agreement.} Krippendorff's $\alpha$ \citep{krippendorff2004content} on the ordinal 1--5 scale across the three annotators is $0.670$ --- by \citeauthor{krippendorff2004content}'s own convention this supports only tentative conclusions about individual-annotator reliability (the customary threshold for confident conclusions is $\alpha \ge 0.800$), and we do not claim more than that here. What licenses the stronger claim below is not this pairwise-style statistic but the three-annotator mean: the intraclass correlation, ICC(A,$k$), is $0.863$, indicating that averaging across annotators yields a reliable composite human score even though single-annotator agreement (reflected in both $\alpha$ and pairwise Spearman $\rho$ between $0.60$ and $0.75$) is more modest.

\begin{figure}[t]
\centering
\includegraphics[width=\columnwidth]{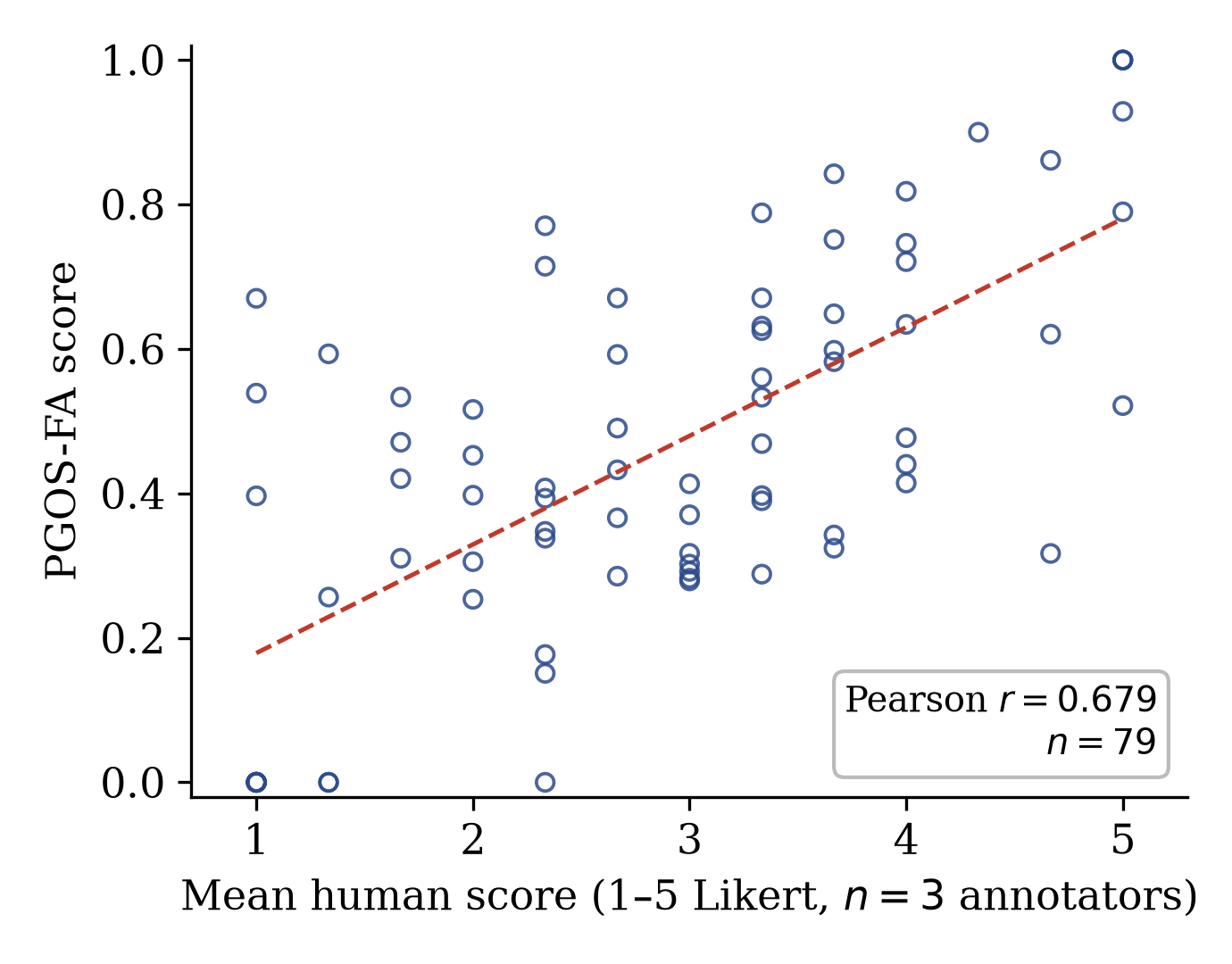}
\caption{PGOS-FA score vs.\ mean human rating for all 79 sampled instances, with a linear trend line. The relationship is positive and reasonably tight given that PGOS-FA is a fully automatic metric being compared against the average of only three human raters.}
\label{fig:humancorr}
\end{figure}

\textbf{Correlation with automatic metrics.} Against this mean human score, PGOS-FA achieves Pearson $r = 0.679$ and Spearman $\rho = 0.623$ (both $n=79$, $p < 10^{-9}$; Figure~\ref{fig:humancorr}); the LLM-judge's binary verdict achieves a point-biserial correlation of $r = 0.584$ ($p \approx 1.6\times10^{-8}$). PGOS-FA therefore tracks human judgment numerically more closely than the LLM-judge does, directly supporting the construct-validity claim that motivated introducing it (Section~\ref{sec:pgosfa}) and complementing Finding~3's statistical-power argument with independent evidence from real human ratings, rather than only from the 16-instance controlled set: on both the artificial controlled set and this sample of real model outputs, PGOS-FA is the more human-aligned of the two automatic metrics studied in this paper.

\subsection{Generalization to a Second Planner Model}\label{sec:generalization}

The ablation (Sections~\ref{sec:method-pipeline}--\ref{sec:significance}) uses a single open-weight backbone (Gemma-3-27B-IT) throughout. To test whether the pipeline's benefit is backbone-specific, we re-ran the best PGOS-FA configuration (RAG + Disambiguation + CoT, English prompt) on the full $n=99$ held-out set with the final planning call substituted for a substantially stronger closed model (Gemini 2.5 Pro), reusing the same persisted retrieval index so retrieval itself is held identical.

\textbf{Isolating the planner as a single variable.} \texttt{gemma-3-27b-it}, the model used for every role in the original pipeline, is no longer served on any endpoint available to us (Section~\ref{sec:limitations}), so the disambiguation and instruction-merger agents in this experiment necessarily run on \texttt{gemma-4-31b-it}, a newer generation from the same open-weight family. Comparing the substituted-planner row directly against an unchanged \texttt{gemma-3-27b-it} row would therefore confound two variables at once: the intended manipulation (the planner) and an incidental one (the disambiguation agents' model generation). We avoid this by reporting three conditions rather than two, adding a matched-generation control in which \texttt{gemma-4-31b-it} serves \emph{every} agent role, including the planner. Because this control row and the Gemini-substituted-planner row then share an identical disambiguation/merger model, the only difference between them is the planner itself. Table~\ref{tab:generalization} reports all three conditions.

\begin{table}[t]
\centering
\caption{Planner backbone comparison, identical pipeline and retrieval index throughout, on the $n=95$ test instances scored successfully under all three conditions (4 excluded by the multi-dependent-verb scoring failure described in Section~\ref{sec:pgosfa}, unrelated to any model substitution). Row 1: Gemma-3-27B-IT in every agent role (original backbone). Row 2: Gemma-4-31B-IT in every role (matched-generation control). Row 3: Gemini~2.5~Pro as the planner, with Gemma-4-31B-IT handling disambiguation/merging.}\label{tab:generalization}
\begin{tabular}{@{}lc@{}}
\toprule
Configuration & Mean PGOS-FA \\
\midrule
Gemma-3-27B-IT (original) & 0.521 \\
Gemma-4-31B-IT (control) & 0.638 \\
Gemini 2.5 Pro (planner) & \textbf{0.678} \\
\botrule
\end{tabular}
\end{table}

Comparing the three rows this way separates the total gain into two distinct, individually testable effects. \textbf{(a) Upgrading the open-weight backbone's generation alone} (row 1 $\to$ row 2, both using Gemma exclusively) already recovers most of the total gain: mean PGOS-FA rises from 0.521 to 0.638 (paired Wilcoxon, $n=95$: $p=0.0003$; \texttt{gemma-4-31b-it} scores higher on 58.9\% of instances). \textbf{(b) With the Gemma generation held fixed at \texttt{gemma-4-31b-it} in both arms}, substituting Gemini 2.5 Pro for the planner produces a further, cleanly isolated gain: 0.638 $\to$ 0.678 ($n=95$: $p=0.017$). This second effect is real but visibly smaller than effect (a), and its win rate is close to 50\% (50.5\%, with 23\% ties) -- the Wilcoxon test is significant because Gemini's wins tend to be larger in magnitude than its losses, not because it wins more often. Section~\ref{sec:externalbaseline} reports an additional, more conservative significance check on this second effect specifically.

Two conclusions follow. First, the pipeline's benefit generalizes beyond the specific backbone used throughout the ablation: a stronger planner produces a further, statistically significant gain on the identical retrieval/disambiguation/prompting setup. Second, of the total 0.157-point gain from row 1 to row 3, roughly three-quarters (0.117 points) is attributable to a newer generation of the \emph{same open-weight model family} used throughout the ablation, and only the remaining quarter (0.039 points) to the specific choice of a stronger, closed, commercial model. Practically, this means meaningful gains on this benchmark are available simply by keeping pace with open-weight model releases, without necessarily requiring paid API access to a closed model at all. We surface a concrete qualitative instance in Section~\ref{sec:qualitative}: a case where both the matched-generation-control and Gemini-substituted planners correctly identify an inverted bowl as a prerequisite obstacle that the original Gemma-3-27B-IT run missed entirely -- itself an illustration that the improvement is visible already at the open-weight generational step.

\textbf{Does the ranking of all eight configurations hold under a different backbone?} The comparison above uses only the single best-performing configuration. To test rank stability directly, we re-ran \emph{all eight} configurations (Table~\ref{tab:configurations}) on \texttt{gemma-4-31b-it} in every agent role, on the identical $n=39$ validation set and identical per-configuration retrieval indices used to produce Table~\ref{tab:ablation}'s English-prompt column. Table~\ref{tab:rankstability} reports both backbones side by side.

\begin{table*}[t]
\centering
\caption{All eight configurations, English prompt, on the $n=39$ validation set: original backbone vs.\ Gemma-4-31B-IT. Ranks are within-backbone (1 = lowest).}\label{tab:rankstability}
\begin{tabular}{@{}lcccc@{}}
\toprule
Configuration & PGOS-FA (Gemma-3) & Rank & PGOS-FA (Gemma-4) & Rank \\
\midrule
1. Baseline & 0.341 & 1 & 0.418 & 1 \\
2. RAG (1-shot) & 0.516 & 3 & 0.665 & 5 \\
3. RAG (3-shot) & 0.535 & 5 & 0.672 & 6 \\
4. RAG (3-shot) + CoT & 0.524 & 4 & 0.649 & 2 \\
5. RAG (unclarified) + CoT & 0.491 & 2 & \textbf{0.694} & \textbf{8} \\
6. RAG + Disambiguation & 0.554 & 7 & 0.661 & 4 \\
7. RAG + Disambiguation + CoT & 0.548 & 6 & 0.677 & 7 \\
8. RAG (CoT ex.) + Disambiguation + CoT & 0.563 & 8 & 0.650 & 3 \\
\botrule
\end{tabular}
\end{table*}

\begin{figure}[t]
\centering
\includegraphics[width=\columnwidth]{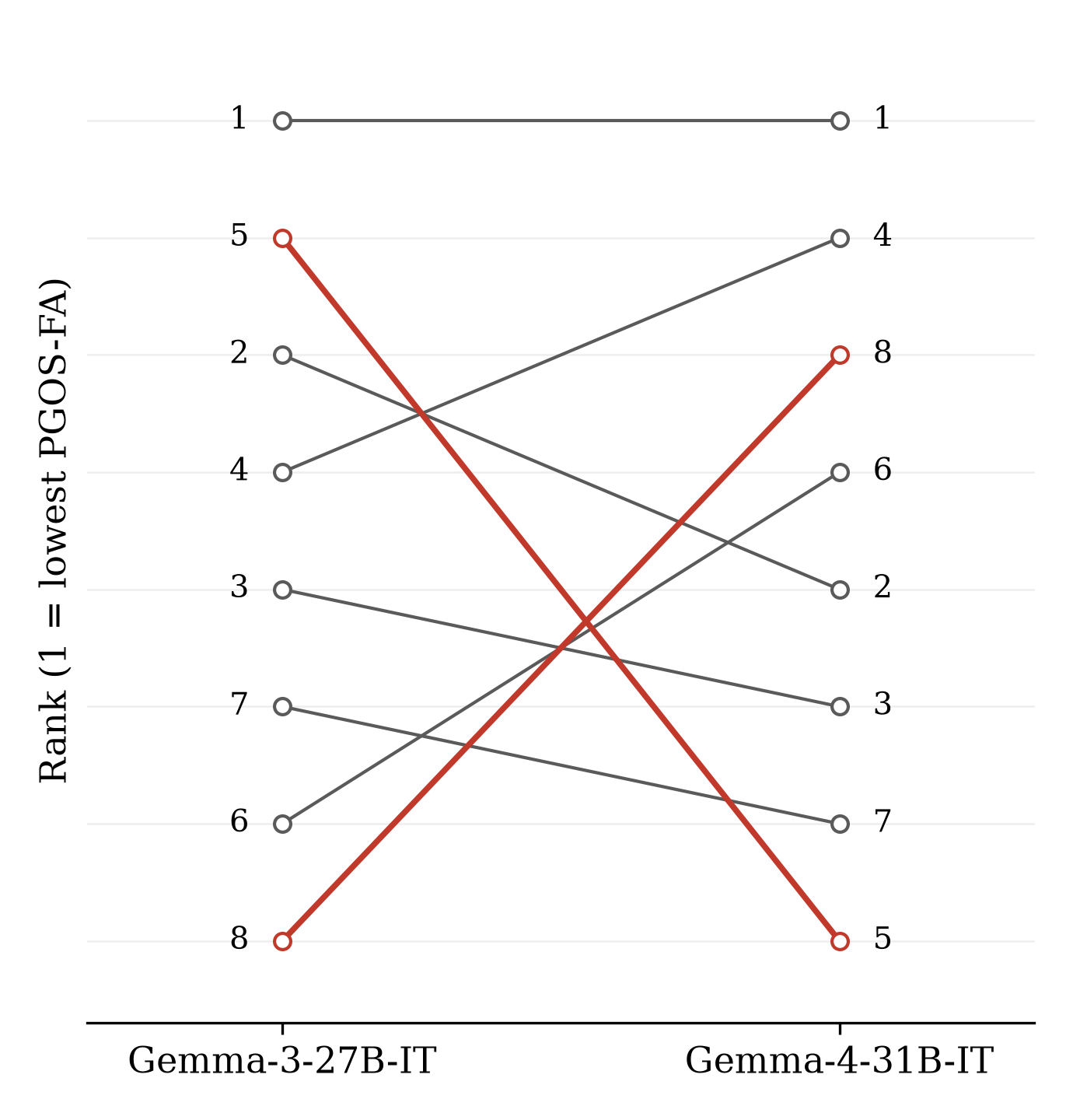}
\caption{Rank of each configuration under Gemma-3-27B-IT (left) vs.\ Gemma-4-31B-IT (right), $n=39$ validation instances. Lines connect the same configuration across backbones; the two most extreme reorderings (configurations 5 and 8) are highlighted. Only configuration 1 (baseline) keeps its rank.}
\label{fig:rankslope}
\end{figure}

Spearman rank correlation across the eight configurations is $\rho = 0.048$ ($p = 0.911$); Kendall's $\tau = 0.000$ ($p = 1.000$) --- essentially uncorrelated, not merely underpowered. What holds is that the baseline remains the worst configuration under both backbones, reinforcing Finding~1. What does not hold is the fine-grained ranking among the seven retrieval-augmented configurations: Figure~\ref{fig:rankslope} shows two configurations moving by five or more ranks between backbones, most notably configuration~5 (rank~2 under Gemma-3) becoming the single best configuration under Gemma-4.

We treat this as a substantive finding in its own right rather than an incidental caveat. Combined with Finding~2's sample-level result, it gives two independent lines of evidence that fine-grained ablation rankings at this scale are not stable properties of the task, while the coarse-grained value of retrieval augmentation is. This distinction --- between robust, well-powered effects and fragile, configuration-specific rankings --- is, in our view, more useful to a practitioner deciding how to build on this benchmark than a single leaderboard-style ranking would be.

\subsection{Comparison Against an External Baseline}\label{sec:externalbaseline}

Every comparison so far is internal to our own ablation. To situate our results against independently published prior work, we reproduce the zero-shot 3-agent pipeline of \citet{brienza2024multiagent} --- the source of the PG2S metric that PGOS-FA extends (Section~\ref{sec:pgosfa}) --- as closely as current model availability allows, using their publicly released prompts and code.\footnote{\url{https://github.com/Lab-RoCoCo-Sapienza/map-vlm}, inspected 2026-09-05; a copy of the relevant files is kept alongside our own code for reproducibility.} Their architecture uses three agents with no retrieval, no few-shot examples, and no chain-of-thought or disambiguation step: a Semantic-Knowledge Miner and a Grounded-Knowledge Miner (both GPT-4o in the original paper) extract, respectively, object-relation triples and a grounded scene description from the image, which a Planner agent (GPT-4-turbo in the original paper) combines into the final plan.

We run this pipeline, prompts unmodified, on the identical 99 held-out test instances used throughout this paper. Two adaptations were necessary and are reported in full: (1)~\texttt{gpt-4-turbo} is no longer served on our available endpoint, so we use \texttt{gpt-4o} for the Planner as well, i.e.\ all three agents share one current GPT-4-class model rather than introducing an additional, less-related substitute; (2)~piloting revealed that, with the prompts used completely verbatim, the Planner answers in English despite receiving a Persian instruction (it has no Persian few-shot signal anywhere in its context, unlike our own configurations), which cannot be meaningfully scored against a Persian reference plan by PGOS-FA's Persian-specific tooling. We therefore append a single instruction to the Planner's system prompt asking it to answer in Persian --- the only content change made to the original prompts, dictated by cross-lingual evaluability rather than anything that could advantage our own method. No retrieval, few-shot exemplars, or reference-plan text are ever included in any of the three agents' prompts; we additionally verified this in code with a runtime assertion checked on every instance.

\begin{table}[t]
\centering
\caption{External baseline comparison, $n=96$ (3 instances excluded by the same scoring failure as Table~\ref{tab:generalization}). Means are computed on these 96 shared instances and so differ slightly from the $n=99$ figures in Table~\ref{tab:testset}. ``External baseline'' is the zero-shot 3-agent GPT-4o pipeline of \citet{brienza2024multiagent}; ``our best configuration'' is RAG + Disambiguation + CoT on Gemma-3-27B-IT.}\label{tab:externalbaseline}
\begin{tabular}{@{}lc@{}}
\toprule
Method & Mean PGOS-FA \\
\midrule
Our baseline (no RAG) & 0.373 \\
External baseline & 0.413 \\
Our best configuration & \textbf{0.531} \\
\botrule
\end{tabular}
\end{table}

Paired Wilcoxon signed-rank tests on the shared 96 instances: our baseline vs.\ the external baseline, $p=0.0141$ (external baseline wins on 55.2\% of instances); external baseline vs.\ our best configuration, $p=0.00003$ (our configuration wins on 66.7\% of instances).

Two conclusions follow. First, our simple baseline is not a strawman: an independently published, architecturally more elaborate method using a stronger commercial model (GPT-4o vs.\ our Gemma-3-27B-IT backbone) only modestly and narrowly outperforms it. Second, and more important for this paper's central claim, our best configuration significantly outperforms this external baseline \emph{despite using a weaker backbone}, indicating the advantage comes from the pipeline design (retrieval from our native-language resource, disambiguation, chain-of-thought) rather than from model capability. We note one honest caveat: the external baseline lacks access to our retrieval index and native Persian training data by design (it is a general-purpose zero-shot method, not built for this dataset), so this is not a controlled ablation of one factor -- it is a comparison against independently developed prior art, which is precisely what it is intended to be.

\textbf{An additional, maximally conservative significance check.} Sections~\ref{sec:significance}, \ref{sec:generalization} and this section together report 40 confirmatory PGOS-FA comparisons: the ablation's 36 (each corrected within its own family, Section~\ref{sec:significance}) plus the generalization section's 2 and this section's 2, each treated above as a single pre-specified test. As a further robustness check, we also pooled all 40 raw $p$-values into one Holm--Bonferroni correction -- the most conservative treatment possible, since it assumes every comparison in the paper competes for the same error budget. Under this pooled correction, Finding~1's 14 vs.-baseline comparisons, the Gemma-3-to-Gemma-4 generational gain ($p=0.0003$), and the external-baseline comparison against our best configuration ($p=0.00003$) remain significant; the Gemma-4-to-Gemini planner gain ($p=0.017$) and the our-baseline-vs-external-baseline comparison ($p=0.0141$) do not. This does not change this paper's central claims: both of the comparisons that fail the pooled check are secondary refinements (how much of the generalization gain the closed model contributes beyond the open-weight upgrade; whether our unmodified baseline is itself distinguishable from prior art) rather than the primary result of their respective sections, both of which remain significant even under this maximally conservative standard. We report this pooled check in the same spirit as Section~\ref{sec:significance}'s Implication: as a transparency exercise rather than a result we would have preferred to hide.

\subsection{Qualitative Analysis}\label{sec:qualitative}

We inspected the 12 lowest-PGOS-FA instances from the best held-out-set configuration (RAG + Disambiguation + CoT, English prompt). A clear, evidence-grounded failure taxonomy emerges.

\textbf{(A) Disambiguation stalls on ``don't know,'' and the plan silently proceeds on the still-ambiguous instruction.} This is the single largest driver of near-zero scores in this sample (5 of 12 cases). The disambiguation agent asks a reasonable clarifying question (e.g., ``there are two keyboards in the image, which one?''), but the automated human-simulator (Section~\ref{sec:method-disambiguation}) cannot answer because the instance's ground-truth clarification Q\&A does not cover that specific question, and returns ``don't know.'' After the maximum three rounds, the planner proceeds on the original ambiguous instruction and picks an arbitrary referent, which then mismatches the reference plan's referent. This is substantially an artifact of automated evaluation, not necessarily of the model's core planning ability --- a human user, unlike the human-simulator, could always answer their own clarifying question.

\textbf{(B) Missed physical/spatial infeasibility.} The model proceeds with a plan despite a physical constraint that the ground truth correctly identifies as blocking (3 of 12 cases, e.g., a storage box too small for all listed items, a bag too small for a laptop). This is the failure mode most directly relevant to the physical common-sense gap discussed for LLM-based planners generally; the generalization experiment (Section~\ref{sec:generalization}) shows this gap narrows, but does not close, with a stronger backbone.

\textbf{(C) False infeasibility (over-rejection).} The inverse of (B): the model declares an instruction not executable when it is (2 of 12 cases) --- in one case because the disambiguation loop exhausted without resolving which of two candidate objects was meant, and the planner defaulted to refusal rather than a reasonable guess.

\textbf{(D) Reference-target mismatch under strict automatic scoring.} In at least one case, the generated plan is arguably a reasonable reading of the instruction but is phrased around a different, also visually plausible, target object than the one the reference plan assumes, driving the automatic score to near-zero despite the output not being obviously wrong. This is a reminder that PGOS-FA (like PG2S before it) scores agreement with \emph{one} reference plan, not plan quality in an absolute sense.

\begin{figure*}[t]
\centering
\begin{minipage}[c]{0.28\textwidth}
\centering
\includegraphics[width=\textwidth]{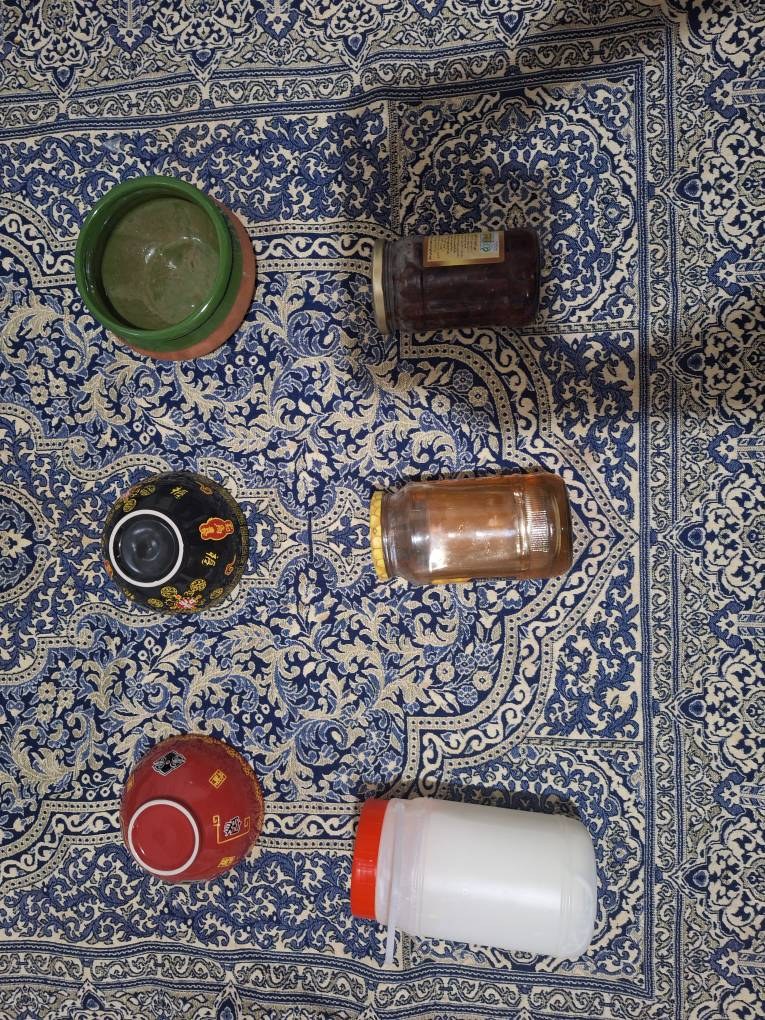}
\end{minipage}
\hfill
\begin{minipage}[c]{0.68\textwidth}
\centering
\includegraphics[width=\textwidth]{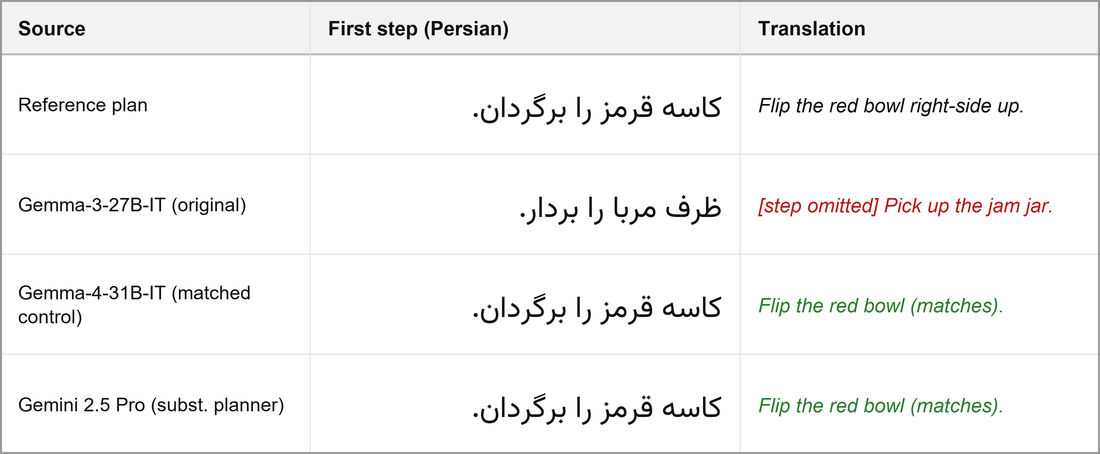}
\end{minipage}
\caption{The red bowl rests upside down in the scene (bottom left of the photograph). The table gives the original Persian first step of each planner's generated plan, with an English gloss. Only the two generalization-experiment planners (Section~\ref{sec:generalization}) recognize the inverted bowl as a prerequisite obstacle; the original Gemma-3-27B-IT run omits this step and proceeds directly to pouring.}
\label{fig:bowl}
\end{figure*}

\textbf{(E) A concrete backbone-dependent miss.} One instance (``pour jam into the red bowl and yogurt into the green pottery bowl'') has a reference plan whose first step is to flip the red bowl right-side up before pouring, since the scene shows it resting upside down (Figure~\ref{fig:bowl}). The original Gemma-3-27B-IT run for this configuration omits this step entirely, pouring directly into the (still inverted) bowl; both the Gemma-4-31B-IT and Gemini 2.5 Pro planners from the generalization experiment (Section~\ref{sec:generalization}) correctly include it, consistent with that section's finding that most of the generalization gain traces to the open-weight generational upgrade rather than to the closed model specifically.

\FloatBarrier

\section{Limitations}\label{sec:limitations}

\textbf{Dataset scale.} At 258 total instances (99 held out for test, 159 for training/retrieval), this resource is small relative to typical NLP benchmark scale, and --- as Section~\ref{sec:significance} shows directly --- this scale is the binding constraint on what claims the ablation can support: only the single largest effect (retrieval vs.\ no retrieval) reaches statistical significance; finer-grained component contributions (disambiguation, CoT, prompt language) are directionally consistent but not distinguishable from noise at $n=39$. We do not know whether these components are truly ineffective or simply under-powered to detect at this scale, and we deliberately avoid claiming the latter is resolved.

\textbf{Ground-truth annotation is single-annotator per instance.} Reference plans and ambiguity/infeasibility labels were produced by the contributing annotator without independent second-annotator verification at collection time.

\textbf{LLM-judge statistical power.} Section~\ref{sec:significance}'s Finding 3 is itself a limitation of one of our two evaluation instruments: the LLM-judge, as deployed here, cannot reliably distinguish configurations at $n=99$. Any claim in this paper about accuracy differences (as opposed to PGOS-FA differences) should be read with this in mind.

\textbf{Single-backbone ablation, generalization checked but not on the LLM-judge metric.} The full eight-configuration ablation uses one open-weight backbone. Section~\ref{sec:generalization} tests generalization to two further backbones in two ways. First, the single best configuration is re-run with the planner substituted, using a matched-generation control to isolate the planner as a single variable; this shows a further, statistically significant gain. Second, all eight configurations are re-run under \texttt{gemma-4-31b-it} to test rank stability directly: the coarse conclusion --- retrieval beats no retrieval --- holds exactly, while the fine-grained ranking among retrieval-augmented configurations does not (Spearman $\rho = 0.048$, essentially uncorrelated). Neither generalization check was scored on the LLM-judge metric, only PGOS-FA, to control cost.

\textbf{Single-image, static-scene grounding.} Unlike video- or simulator-grounded benchmarks (EgoCOT, ALFRED), our task and dataset do not capture temporal state changes during execution; a generated plan is judged against a single before-state image, not verified by actually executing it. This trades execution-grounded verification for annotation tractability and dataset scale, but means correctness here is a proxy for, not a guarantee of, real-world executability.

\textbf{Language and cultural scope.} The dataset targets Persian instructions in Iranian home/office settings specifically; transfer of either the dataset's specific findings or PGOS-FA's tooling choices (Tooka-SBERT, Persian dependency parsing) to other low-resource languages is untested and would require new language-specific resources.

\section{Conclusion}\label{sec:conclusion}

We presented the first Persian vision-language-action dataset explicitly designed around two under-studied failure modes of instruction-following: ambiguity and physical infeasibility. Alongside it, we conducted a rigorous empirical study of retrieval, chain-of-thought, and disambiguation techniques on this resource, and introduced a new evaluation metric, PGOS-FA, shown to be more statistically reliable than LLM-as-judge scoring on identical data. Our statistical analysis intentionally reports what the data can and cannot yet support: retrieval-augmentation's benefit is unambiguous, while several finer-grained design choices remain promising but under-powered to confirm at the current dataset scale, and further gains on the identical pipeline are available from both a newer open-weight model generation and, to a smaller additional degree, a substantially stronger closed planner model. We see this combination --- a richly annotated low-resource benchmark, available to other researchers on request, paired with an honest accounting of statistical power --- as a more durable foundation for future work than a narrower claim of a single best-performing method, and we hope it lowers the barrier to studying embodied instruction-following beyond English.

\section*{Supplementary Information}

The dataset, PGOS-FA implementation, all pipeline prompts, and the analysis scripts behind every table in this paper will be made available from the corresponding author upon reasonable request following publication.

\section*{Declarations}

\begin{itemize}
\item \textbf{Funding:} Not applicable -- this research received no external funding.
\item \textbf{Conflict of interest/Competing interests:} The authors declare no competing interests.
\item \textbf{Ethics approval:} Not applicable (the study involved no experimentation on live vertebrates, humans, or human tissue samples in the sense requiring formal institutional ethics review; the 19 dataset contributors' involvement was limited to photographing household/office scenes and writing instructions, and the 3 additional annotators in the human-correlation study (Section~\ref{sec:humancorr}) rated existing model outputs on a Likert scale, with informed consent obtained from all participants as described below).
\item \textbf{Consent to participate:} Informed consent was obtained from all 19 data-collection contributors for their contributions to be used and released as part of this research dataset, and from all 3 annotators in the human-correlation study (Section~\ref{sec:humancorr}) for their ratings to be used in this research.
\item \textbf{Consent for publication:} Not applicable (no identifiable personal data of third parties is included; dataset photographs depict household/office objects, not people).
\item \textbf{Availability of data and materials:} The dataset is available from the corresponding author upon reasonable request following publication.
\item \textbf{Code availability:} All code (pipeline, PGOS-FA implementation, statistical analysis scripts) is available from the corresponding author upon reasonable request following publication.
\item \textbf{Authors' contributions:} A.P.\ conceived the study, collected and curated the dataset, and implemented the pipeline and evaluation tooling. M.R.\ wrote the manuscript and conducted the supplementary experiments (generalization, external-baseline, and significance analyses). S.S.S.\ supervised the research. All authors reviewed and approved the final manuscript.
\end{itemize}

\bibliographystyle{apalike}
\bibliography{sn-bibliography}

\end{document}